\pdfoutput=1

\documentclass[11pt]{article}

\usepackage{emnlp2021}

\usepackage{times}
\usepackage{latexsym}

\usepackage[T1]{fontenc}

\usepackage[utf8]{inputenc}

\usepackage{microtype}

\usepackage[ruled,linesnumbered]{algorithm2e}
\usepackage{algpseudocode}
\usepackage{amsmath}
\usepackage{times}
\usepackage{latexsym}
\usepackage{booktabs}
\usepackage{multirow}
\usepackage{graphicx}

\usepackage{color}
\usepackage{booktabs}

\usepackage{bm}
\usepackage{amsfonts}

\usepackage{xcolor}
\usepackage{color, soul}

\newcommand{\nop}[1]{}

\title{Learning from Multiple Noisy Augmented Data Sets for Better Cross-Lingual Spoken Language Understanding}

\author{
  {Yingmei Guo}$^1$\footnotemark[1] \quad {Linjun Shou}$^2$ \quad {Jian Pei}$^3$ \quad \textbf{Ming Gong}$^2$ \quad {Mingxing Xu}$^1$ \quad \\ \textbf{Zhiyong Wu}$^1$ \quad \textbf{Daxin Jiang}$^2$\footnotemark[2] \\
  $^1${\small Department of Computer Science and Technology, Tsinghua University} \\
  $^2${\small NLP Group, Microsoft STCA} \\
  $^3${\small School of Computing Science, Simon Fraser University} \\
  \texttt{\small \{guoym18@mails.,xumx@,zywu@sz.\}tsinghua.edu.cn}\\
  \texttt{\small \{lisho,migon,djiang\}@microsoft.com} 
  \texttt{\small jpei@cs.sfu.ca} \\
}

\date{}

\begin{document}
\maketitle

\footnotetext[1]{Work is done during internship at NLP Group, Microsoft STCA.} 
\footnotetext[2]{Corresponding author.} 

\begin{abstract}

Lack of training data presents a grand challenge to scaling out spoken language understanding (SLU) to low-resource languages. Although various data augmentation approaches have been proposed to synthesize training data in low-resource target languages, the augmented data sets are often noisy, and thus impede the performance of SLU models. In this paper we focus on mitigating noise in augmented data. We develop a denoising training approach. Multiple models are trained with data produced by various augmented methods. Those models provide supervision signals to each other. The experimental results show that our method outperforms the existing state of the art by 3.05 and 4.24 percentage points on two benchmark datasets, respectively. The code will be made open sourced on github.

\end{abstract}

\section{Introduction}\label{sec:intro}

Spoken language understanding (SLU) is a key component in task-oriented dialogue systems. SLU consists of two subtasks: intent detection and slot tagging~\citep{WANG2005,TUR2011}. Although promising progress has been achieved on SLU in English~\citep{LIU2016,peng2020data,huang-etal-2020-federated}, those methods need large amounts of training data, and thus cannot be applied to low-resource languages where zero or few training data is available. 

In this paper, we target at the extreme setting for cross-lingual SLU where no labeled data in target languages is assumed, which is critical for industry practice, since annotating a large SLU dataset with high quality for every language is simply infeasible.

Existing cross-lingual transfer learning methods mainly build on pre-trained cross-lingual word embeddings~\citep{ruder2019survey} or contextual models~\citep{wu2019beto,huang2019unicoder,lample2019cross,conneau2019unsupervised}, which represent texts with similar meaning in different languages close to each other in a shared vector space. Those approaches often show good performance on intent detection. The results on slot tagging, however, are often unsatisfactory, especially for distant languages, which are dramatically different from English in scripts, morphology, or syntax~\cite{UPADHYAY2018,schuster2018cross,li2020mtop}.

Several studies~\citep{schuster2018cross,liu2019attention,li2020mtop} show that adding translated data into the fine-tuning process of pretrained models can improve the results of cross-lingual SLU substantially when no golden-labeled training data in target languages is available. For example, machine translation can be employed to translate the training data in English into target languages, and some alignment methods, such as attention weights~\citep{schuster2018cross}, fastalign~\citep{dyer2013simple} or giza++~\citep{och03:asc}, can be further applied to label the translated data. Another approach to alleviate the problem of data scarcity is to automatically generate training data. Recently, some methods for monolingual SLU~\citep{anaby2020not,kumar2020data,zhao2019data, peng2020data} automatically label domain-specific data or use pre-trained language models to generate additional data. 

However, the synthesized training data derived from both the translation approach and the data generation approach may be quite noisy and may contain errors in label. 
For the translation approach, both the translation process and the alignment process may generate errors~\cite{XU2020,LI2020a}.
For the data generation approach, it is often hard to control a right tradeoff between generating correct but less diverse data and generalizing more diverse data but with more noise. Moreover, generating synthetic training data across languages further adds challenges to the robustness of the generation methods.

\nop{
\begin{figure}[t] 
\setlength{\abovecaptionskip}{3pt} \setlength{\belowcaptionskip}{-10pt}
\centering 
\includegraphics[width=0.42\textwidth]{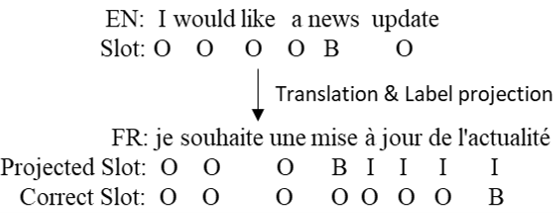} 
\caption{Noise introduced by the alignment process. B denotes B-NEW\_TYPE, I denotes I-NEWS\_TYPE, O denotes NO\_LABEL.} 
\label{fig:sen}
\end{figure}
}

To filter out noise in the synthesized training data, a few methods are proposed, such as the mask mechanism~\citep{li2020mtop}, the soft alignment method~\citep{XU2020}, the unsupervised adaptation method~\citep{LI2020a}, the rule-based filtering method~\citep{peng2020data}, the classifier based filtering method~\citep{anaby2020not} and the language model score based filtering method~\citep{shakeri2020multilingual}. These methods rely on either ad-hoc rules or extra models. Although they have shown promising results, each of them considers only a single source for data augmentation. It is still challenging to differentiate noisy instances from useful ones, since all those instances are sampled from the same distribution generated by the same method.

In this paper, we regard both the translation approach and the generation approach as data augmentation methods. We tackle the problem of reducing the impact of noise in augmented data sets. We develop a principled method to learn from multiple noisy augmented data sets for cross-lingual SLU, where no golden-labeled target language data exists. Our major technical contribution consists of a series of denoising methods including \emph{instance relabeling}, \emph{co-training} and \emph{instance re-weighting}.

First, motivated by the self-learning method~\citep{zoph2020rethinking}, we design a model-ensemble-based instance relabeling approach to correct the noisy labels of augmented training data in low resource languages. As teacher models can not always generate correct labels, the original self-learning method tends to suffer from accumulated errors caused by model predictions. To alleviate the problem of accumulated errors, in our instance relabeling approach, we use crowd-intelligence from multiple models to derive the more reliable labels of pseudo training instances. Besides, we filter out noisy instances based on co-training and instance re-weighting strategies to reduce the impact of incorrect predictions on subsequent training. Our training strategy does not follow a traditional teacher-student manner. Instead, we use model predictions in the last epoch as pseudo labels in the current epoch to compute loss which saves training time.

In order to filter out noisy instances, we adopt a co-training mechanism, which uses selected instances from the other models to train the current model. Different from the co-teaching method~\citep{han2018co} where two models are trained with the same data, we propose multiple models should be trained with multiple different noisy augmented data where noise may be largely independent. It is because deep neural networks have a high capacity to fit noisy labels. When two models are trained with the same data, we tend to obtain two similar models. The co-teaching method gradually becomes a naive selection method with two models ensembled. In our co-training method, by employing very different data generation methods, we hope to attain that the noise from different sources may be largely independent and models can learn different knowledge from them. Therefore, the instances that pass the screening process of the other models can serve as the supervision signals to the current model which alleviates the problem of accumulated errors caused by selection bias.

Last, we further propose an instance re-weighting technique to adjust the weights of training instances adaptively. As we do not have real training data in target languages, we can use the consistency among the soft labels predicted by different models to predict the reliability of the instances. Intuitively, if the predictions of different models are highly inconsistent on an instance, the instance may contain much noise. The larger deviation, the more uncertainty, and the less weight. This idea further increases the robustness of the selected training instances.

We conduct extensive experiments on two public datasets. The experiment results clearly indicate that, by consciously considering multiple noisy data sources derived from very different augmentation methods, our approach is more effective than using any single source. Our methods improve the state of the art (SOTA) by 3.05 and 4.24 percentage points on the two benchmark datasets, respectively. 

\section{Related Work}
\textbf{The cross-lingual spoken language understanding} methods can be divided into two main categories: the model transfer methods and the data transfer methods. 

The model transfer methods build on pre-trained cross-lingual models to learn language agnostic representations, such as MUSE~\citep{lample2017unsupervised}, CoVE~\citep{mccann2017learned}, mBERT~\citep{wu2019beto}, XLM~\citep{lample2019cross}, Unicoder~\citep{huang2019unicoder}, and XLM-R~\citep{conneau2019unsupervised}. The English training data is applied to fine-tune the pre-trained models and then the fine-tuned models are directly applied to target languages~\cite{liu2019attention,schuster2018cross,qin2020cosda,UPADHYAY2018,li2020mtop}. To better align embeddings between source and target languages, \citet{liu2019zero} use domain-related word pairs and employ a latent variable model to cope with the variance of similar sentences across different languages. \citet{liu2019attention} and \citet{qin2020cosda} use parallel word pairs to construct code-switching data for fine-tuning. Their methods encourage the model to align similar words in different languages into the same space and attend to keywords. \citet{liu2020cross} propose a regularization approach to align word-level and sentence-level representations across languages without any external resource. 

The data transfer methods construct pseudo-labeled data in target languages. These methods usually employ machine translators to translate training instances in a source language into target languages and then apply alignment methods, such as attention weights~\citep{schuster2018cross}, fastalign~\citep{dyer2013simple}, or giza++~\citep{och03:asc}, to project slot labels to the target language side. The derived training instances are combined with the training data in the source language to fine-tune the pre-trained cross-lingual models. Previous studies~\cite{UPADHYAY2018,schuster2018cross,li2020mtop} show that adding translated training data can significantly improve the model performance, especially on languages which are distant from the source language.  

The data generation approaches can also construct additional training data. Some methods~\citep{wang2015s,Marivate2020improving,GAO2020} make slight changes to the original training instances through word replacement or paraphrases. More sophisticated methods generate training data through large-scale neural networks, such as generative adversarial networks~\citep{goodfellow2020generative}, variational autoencoders~\citep{doersch2016tutorial,tanaka2019data,russo2020control}, and pre-trained language models~\citep{wu2019conditional,anaby2020not,kumar2020data,peng2020data}.  


\begin{figure*}[t] 
\centering 
\includegraphics[width=0.9\linewidth]{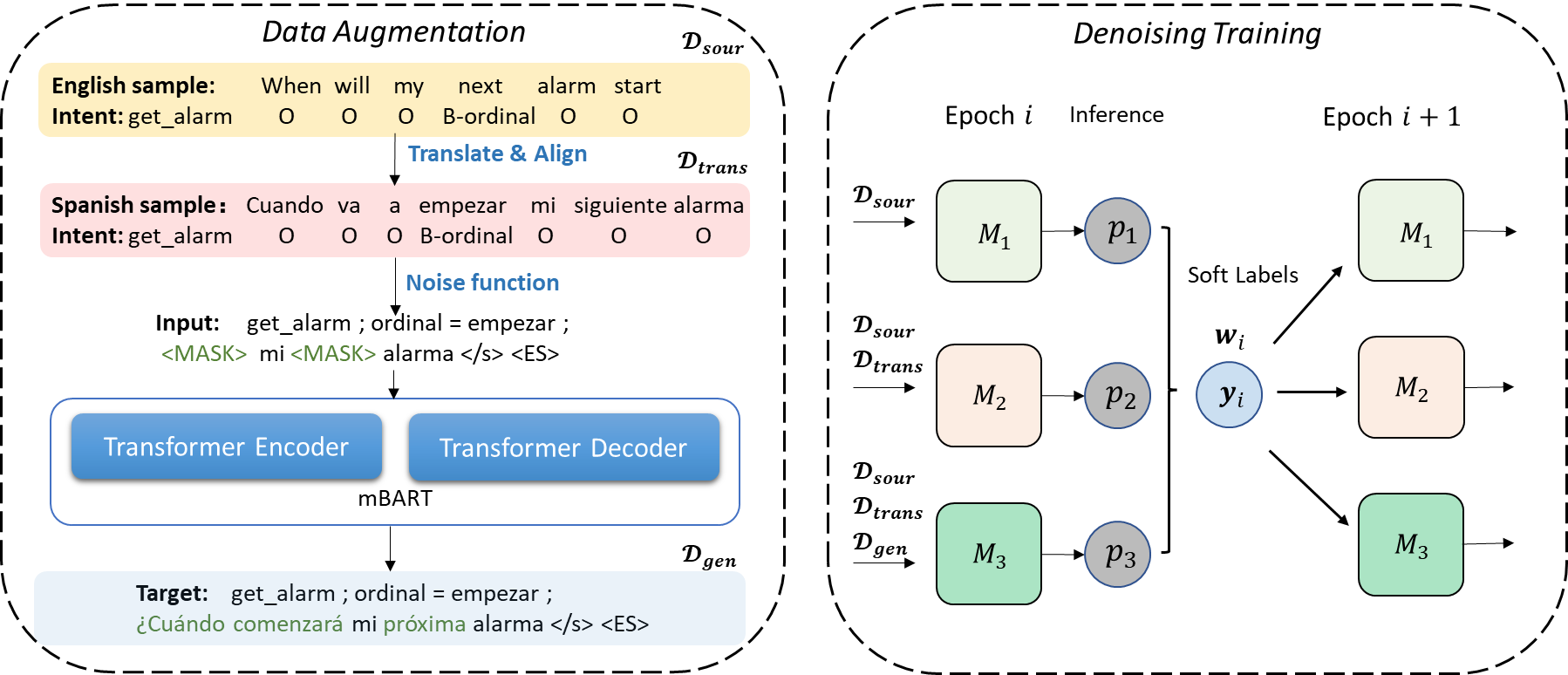}
\caption{The overall framework of data augmentation module (left) and denoising module (right).}
\label{fig:main}
\vspace{-0.3cm}
\end{figure*}

\section{Method}

In this section, we define the problem and then propose our method.

\subsection{Problem Definition and Solution Framework}
The SLU task aims to parse user queries into a  predefined semantic representation format. Formally, given an utterance $\bm{x}={\{x_i\}}^L_{i=1}$ with a sequence of $L$ tokens, a SLU model targets to produce an intent label $\bm{y}^{I}$ for the whole utterance and a sequence of slot labels $\bm{y}^{S} ={\{\bm{y}_i^S\}}_{i=1}^L$, where $\bm{y}_i^S$ is the slot label for the $i^{th}$ token $x_i$. Here, we target at the extreme cross-lingual setting where only some training data $\mathcal{D}_{src}$ in English (or, in general, a rich-resource source language) and some development data $\mathcal{D}_{dev}$ in English exist. Besides, some annotated data $\mathcal{D}_{test}$ in target languages is used as the test set. Cross-lingual SLU is to learn a model by leveraging $\mathcal{D}_{src}$ to perform well on $\mathcal{D}_{test}$, using $\mathcal{D}_{dev}$ for parameter tuning.

We add a special token [CLS] in front of each input sequence.  Then we feed $\bm{x}$ into an encoder $\mathcal{M}_{enc}$ to obtain the contextual hidden representation $\bm{H}={\{\bm{h}_i\}}_{i=0}^L$, that is,
$\bm{H}=\mathcal{M}_{enc}(\bm{x}; \Theta)$,
where $\Theta$ denotes the parameters of the encoder. 

We take $\bm{h}_{0}$ as the sentence representation for intent classification and take $\bm{h}_{i}$ $(1\leq i\leq L)$ as the token representations for slot filling. We apply linear transformation and the \emph{softmax} operation to obtain the intent probability distribution $\bm{p}^{I}(\bm{x}; \Theta)$ and the slot probability distribution $ \bm{p}^{S}_{i}(\bm{x}; \Theta)$, that is,
\begin{equation}
\small
\begin{aligned}
&\bm{p}^{I}(\bm{x}; \Theta) = \emph{softmax} (\bm{W}^{I} \cdot \bm{h}_0+\bm{b}^{I}) \\
& \bm{p}^{S}_{i}(\bm{x}; \Theta) = \emph{softmax} (\bm{W}^{S} \cdot \bm{h}_i+\bm{b}^{S})
\end{aligned}
\end{equation}
where $\bm{p}^{I} \in \mathbb{R}^{1 \times \left| C^{I} \right|}$, $\bm{p}^{S}_{i} \in \mathbb{R}^{1 \times \left| C^{S} \right|}$, $C^{I}$ is the set of intent labels, $C^{S}$ is the set of slot labels under the BIO annotation schema~\cite{ramshaw1999text}, $\bm{W}^I \in \mathbb{R}^{\left| C^{I} \right| \times d}$ and $\bm{W}^S \in \mathbb{R}^{\left| C^{S} \right| \times d}$ are the output matrices, and $\bm{b}^{I}$ and $\bm{b}^{S}$ are the biases.

The overall architecture of our proposed method is shown in Figure~\ref{fig:main}. It consists of two major modules, the data augmentation module and the denoising module. 

\subsection{The Data Augmentation Module}
\label{sec:augmentation}

In this module, we augment training data in target languages via translation and generation. The left part of Figure~\ref{fig:main} shows the details. 

\subsubsection{Translation} We use Google Translator to translate the training corpus  $\mathcal{D}_{src}$ in the source language (English) to the target languages. In addition to translation, we also need some word alignment methods to project the slot labels to the target language side. We try giza++~\cite{och03:asc} and fastalign~\cite{dyer2013simple} to obtain word alignment information and find that the pseudo slot labels projected by giza++ generally lead to better performance (about 2\% increase in F1 on the SNIPS dataset). Thus, in the rest of the paper, we use Google translator and giza++ to produce translated data. Denote by $\mathcal{D}_{trans}$ the translated training corpus.

\subsubsection{Generation} 

To further increase the diversity of synthesized training data, we leverage multilingual BART (mBART)~\citep{liu2020multilingual} as the generator to synthesize additional target language training corpus. 
Specifically, we first fine-tune the pretrained mBART model on the translated training data  $\mathcal{D}_{trans}$ by adopting the denoising objective~\citep{liu2020multilingual} - the cross-entropy loss between the decoder’s output and the original input. The input to mBART consists of the dialog act and the utterance in $\mathcal{D}_{trans}$, defined by
 \begin{equation}
     \bm{X} = [I;(s_1,v_1),...,(s_p,v_p);(x_1,...,x_L)]
\label{input}
 \end{equation}
 where $I$ is the intent and ${(s_i,v_i)}_{i=1}^p$ are the slot-value pairs. Here, $v_i$ in the target language is obtained by word alignment between utterances in the source and the target language. Following~\citet{liu2020multilingual}, we apply text infilling as the injected noise in the fine-tuning stage. 

After fine-tuning, we apply the same noise function to the input data $\bm{X}$ and leverage the fine-tuned mBART to generate $m$ candidates for each instance.
To increase diversity of generated data, the top-$p$ sampling strategy~\citep{Fan2018hierarchical} is adopted. Each generated instance consists of ``an utterance, the corresponding intent and slot-value pairs''. Then, we perform preliminary data selection by filtering out the generated utterances not containing the required slot values. Last, we randomly sample $a$ instances from the candidate set for each input to construct the generated corpus $\mathcal{D}_{gen}$.

\subsection{The Denoising Module}
\label{sec:denoising}
To tackle the noisy label problem introduced by the data augmentation module, we design a denoising module shown in Algorithm~\ref{alg}.


At the initialization stage, we first train $K$ models using the augmented data derived from $K$ different augmentation methods. All models are optimized by the cross-entropy loss function computed using the original labels of intent and slots. For the $k$-th model ($k \le K$),
\begin{equation}
\setlength{\abovedisplayskip}{3pt}
\setlength{\belowdisplayskip}{3pt}
\small
   \mathcal{L}^k(\bm{x})\! = -\frac{1}{L}\sum^{L}_{j=1}\bm{y}_{j}^Slog\bm{p}^{S}_{ j}(\bm{x};\Theta_k)\!-\!\bm{y}^{I}log\bm{p}^{I}(\bm{x};\Theta_k)
\label{eq:cross-entropy}
\end{equation}
where $\bm{x}$ is a training utterance, $\bm{y}^{I}$ is the intent label, $\bm{y}_j^S$ is the slot label of the $j$-th word in the utterance, and $\bm{p}^{I}(\bm{x};\Theta_k)$ and $\bm{p}^{S}_j(\bm{x};\Theta_k)$ are the predicted probability distributions of the intent and the slot, respectively.

To keep our discussion simple, in this paper, we mainly consider using the training corpora derived from machine translation and generation. Thus, we maintain $K=2$ SLU networks with the same structure in the training process. $\mathcal{M}_1$ and $\mathcal{M}_2$ are trained using different training corpora $\mathcal{D}_1\!=\!\{\mathcal{D}_{src},\mathcal{D}_{trans}\}$ and $\mathcal{D}_2\!=\!\{\mathcal{D}_{src},\mathcal{D}_{trans},\mathcal{D}_{gen}\}$, respectively.  Based on our epxeriments, $\mathcal{D}_{gen}$ only is too noisy thus we combine with $\mathcal{D}_{trans}$. Our training framework in general can handle more than $2$ models. We present experimental results with more than $2$ models in Section~\ref{sec:analysis}.

After the initialization stage, each model has learned some knowledge from each augmented training data. Since there exists noise in the augmented training data (such as $\mathcal{D}_{trans}$ and $\mathcal{D}_{gen}$ ), we step into the relabeling stage, which combines a series of strategies: \emph{instance relabeling}, \emph{co-training} and \emph{instance re-weighting} to reduce the impact of the noise.

\begin{algorithm}
  \caption{Denoising Module\label{alg}} 
  \small
  \label{alg::denoising}
    \KwData{$\mathcal{M}_k (1 \le k \le K)$: SLU networks;
      $E_{all}$: the number of fine-tuning epochs;
      $E$: the number of initialization epochs;
      $\mathcal{D}_{src}, \mathcal{D}_{trans}, \mathcal{D}_{gen} $: training corpus;} 
    \KwResult{$\mathcal{M}_k (1 \le k \le K)$}
    \tcp{Initialization Stage}
    \For {$e=1$; $e < E$; $e++$}{
        shuffle training corpus $\mathcal{D}_k$ from $\{\mathcal{D}_{src},\mathcal{D}_{trans},\mathcal{D}_{gen},...\}$\\
        update $\mathcal{M}_k$ with $\mathcal{L}^k(\mathcal{D}_k)$\\}
    \tcp{Relabeling Stage}
    \For{$e = E$; $e \le E_{all}$; $e++$} 
      {shuffle training corpus $\mathcal{D}=\{\mathcal{D}_{src}, \mathcal{D}_{trans}, \mathcal{D}_{gen},...\}$ into $B$ mini-batches;\\
      \For{$j=1$; $j<=B$; $j++$}
        {Fetch $j^{th}$ mini-batch $\mathcal{D}^\prime$ from $\mathcal{D}$ \\
        \tcp{Co-Training}
        $\mathcal{D}^\prime_k\!=\!argmin_{\mathcal{D}^\prime:|\mathcal{D}^\prime|\ge(1-\delta)|\mathcal{D}^\prime|}\sum_i^{K\setminus k}\mathcal{L}^i(\mathcal{D}^\prime)$ \\
         \tcp{Instance Re-weighting}
        update $\mathcal{M}_k$ with $\mathcal{L}^{k'}(\mathcal{D}^\prime_k)$ \\
         \tcp{Instance Relabeling}
        update labels of instances in $\mathcal{D} \setminus \mathcal{D}_{src}$ as Equation~\ref{eq:relabel}
        }
        }
\end{algorithm}

\subsubsection{Instance Relabeling}

In the relabeling stage, model training and instance relabeling are conducted iteratively. Motivated by the idea of model ensemble, we use the ensemble of model predictions to correct label errors in a self-learning manner. Specifically, all models are trained using all training corpora $\mathcal{D}=\{\mathcal{D}_{src},\mathcal{D}_{trans},\mathcal{D}_{gen}\}$. The slot labels and the intent labels of the training instances in $\mathcal{D}_{trans}$ and $\mathcal{D}_{gen}$ are modified to the corresponding ensemble predicted probability distributions, which are used as the pseudo-truth labels to compute the loss in the next epoch.  That is,
\begin{equation}
\small
\setlength{\abovedisplayskip}{3pt}
\setlength{\belowdisplayskip}{3pt}
\begin{aligned}
    \widetilde{\bm{y}}^{I} = \frac{1}{K}\sum^{K}_{k=1}\bm{p}^{I}(\bm{x};\Theta_k); 
    \quad  \widetilde{\bm{y}}_j^S = \frac{1}{K}\sum^{K}_{k=1}\bm{p}^{S}_{j}(\bm{x};\Theta_k) 
\end{aligned}
\label{eq:relabel}
\end{equation}

\subsubsection{Co-Training}

Heuristically, instances with small losses are more likely to have cleaner labels.  When noise from different augmentation methods is more or less independent, each model can learn instances with small cross-entropy losses from the other models. 

Specifically, when $K=2$, in each batch of the training data, each network discards the instances with larger losses computed by Equation~\ref{eq:cross-entropy} by a ratio of $\delta$ and then teach the remaining instances to another one.


\subsubsection{Instance Re-weighting}
Another way to reduce the impact of noisy instances is to assign different weights to different instances, the more noisy an instance, the less weight it is associated. We design a re-weighting mechanism to implement this idea. The intuition is that if the predictions by multiple models are quite inconsistent, the instance may likely be noisy.  Technically, we design an uncertainty based weight to re-weight the training instances.  The larger the deviation, the more uncertainty and the less the weight. Specifically, the uncertainty of each instance is defined as: 
\begin{equation}
\setlength{\abovedisplayskip}{3pt}
\setlength{\belowdisplayskip}{3pt}
      u=\frac{1}{K}\sum^K_{k=1}Var(\bm{p}(\bm{x};\Theta_k),\bm{m}(x))
\end{equation}
where
\begin{equation}
\setlength{\abovedisplayskip}{3pt}
\setlength{\belowdisplayskip}{3pt}
\small
\begin{aligned}
     &Var(\bm{p}(\bm{x};\Theta_k),\bm{m}(x))=(\bm{p}^{I}(\bm{x};\Theta_k)-\frac{1}{K}\sum^K_{k=1}\bm{p}^I(\bm{x};\Theta_k))^2 \\
      &+\frac{1}{L}\sum_{l=1}^{L}(\bm{p}^{S}_{l}(\bm{x};\Theta_k)-\frac{1}{K}\sum^K_{k=1}\bm{p}^S_l(\bm{x};\Theta_k))^2
\end{aligned}
\end{equation}

We further compute weight by $w=e^{-u}$ and incorporate this weight into Equation~\ref{eq:cross-entropy} to obtain: 
\begin{equation}
\setlength{\abovedisplayskip}{3pt}
\setlength{\belowdisplayskip}{3pt}
\small
    \mathcal{L}^{k'}(\bm{x})=-w[
    \frac{1}{L}\sum^{L}_{j=1}\widetilde{\bm{y}}_{j}^Slog\bm{p}^{S}_{ j}(\bm{x};\Theta_k)\!+\!\widetilde{\bm{y}}^{I}log\bm{p}^{I}(\bm{x};\Theta_k)
    ]
\label{eq:modi}
\end{equation}
which is the new training objective during the relabeling stage.  

\begin{table*}[h!]
\setlength{\abovecaptionskip}{3pt} 
\setlength{\belowcaptionskip}{-10pt}
\begin{center}
\resizebox{\textwidth}{!}{%
 \begin{tabular}{l c c c c c c c c c c}
 \toprule
   & \multicolumn{3}{c}{\textbf{SNIPS} (F1/Intent Accuracy)} &
   & \multicolumn{6}{c}{\textbf{MTOP} (Exact Match Accuracy)} \\
   & es & th & Average & 
   & es & fr & de & hi & th & Average \\
 \midrule
   Multi. CoVe
   & 19.25/53.89 & 35.62/70.70 & 27.44/62.30 &
   & - & - & - & - & - & - \\
   Transferable Latent Variable
   & 65.79/90.20 & 32.24/73.47 & 49.02/81.84 &
   & - & - & - & - & - & - \\
    Attention-Informed Mixed
   & 73.89/87.88 & 27.12/73.46 & 50.51/80.67 &
   & - & - & - & - & - & - \\
    CoSDA-ML
   & 80.40/94.80 & 37.30/76.80 & 58.85/85.80 &
   & - & - & - & - & - & - \\
    LR\&ALVM 
   & 72.49/92.31 & 33.28/75.77 & 52.89/84.04 &
   & - & - & - & - & - & - \\
 \midrule
   EN
   & 84.20/97.70 & 46.00/90.40 & 65.10/94.05&
   & 69.10 & 65.40 & 64.00 & 55.00 & 43.80 & 59.46 \\
   EN + Trans.
   & 73.40/98.30 & 50.90/96.60 & 62.15/97.45 &
   & 74.50 & 72.60 & 64.70 & 58.30 & 56.50 & 65.32 \\
   EN + Trans. + mask
   & 83.00/98.00 & 50.20/96.60 & 66.60/97.30 &
   & 74.60 & 72.20 & 65.70 & 62.50 & 53.20 & 65.64 \\
   EN\textsubscript{reimp}
   & 84.69/97.67 & 39.34/89.42 & 62.02/93.55 &
   & 68.38 & 70.48 & 66.15 & 54.81 & 38.36 & 59.63  \\
   EN + Trans.\textsubscript{reimp}
   & 83.58/99.34 & 79.29/97.93 & 81.44/98.64 &
   & 70.25 & 67.27 & 70.39 & 64.22 & 55.80 & 65.59 \\
 \midrule
   EN + Trans. + Gen. + Denoise
  & 86.19/99.37 & 82.79/98.17 & \textbf{84.49/98.71} &
   & 71.58 & 71.53 & 74.05 & 69.66 & 62.31 & \textbf{69.83} \\
 \bottomrule
\end{tabular}}
\end{center}
\caption{Comparison results between our approach and prior SOTA methods. The top block rows are baselines that belong to model transfer methods. The middle block rows are XLM-R large model based approaches, w/ or w/o translated training data. Methods with \textsubscript{reimp} are re-implemented in this paper with different translator and alignment tool. EN, Trans. and Gen. denote source language data $\mathcal{D}_{src}$, translated target language data $\mathcal{D}_{trans}$ and generated target language data $\mathcal{D}_{gen}$, respectively. Denoise denotes proposed denoising module.
}
\label{tab:main_table}
\end{table*}

\section{Experiments}
\label{experiments}

In this section, we report our experiments on two benchmark datasets.

\subsection{Settings}

We evaluate the effectiveness of our proposed approach over five languages on two benchmark datasets: \textbf{SNIPS}~\citep{schuster2018cross} and \textbf{MTOP}~\citep{li2020mtop}. The details of datasets are provided in Appendix.

For generation, we generate $m=10$ candidates for each input and randomly sample $a=1$ from each candidate set to construct $\mathcal{D}_{gen}$. Our SLU model is based on the pre-trained XLM-R large model, which has $L=24$ layers and two additional task-specific linear layers for intent classification and slot filling.
More implementation details including hyper-parameters are described in Appendix.

Following the previous works~\citep{schuster2018cross,li2020mtop}, we use F1 score to measure the slot filling quality and use accuracy score to evaluate the intent classification quality on the SNIPS dataset and use Exact Match Accuracy on the MTOP dataset. 

We employ the following SOTA baselines in two groups. The first group is the model transfer methods, including \textbf{Multi.CoVe}~\cite{schuster2018cross}; \textbf{Transferable Latent Variable}~\cite{liu2019zero}; \textbf{Attention-Informed Mixed}~\cite{liu2019attention}; \textbf{CoSDA-ML}~\cite{qin2020cosda}; \textbf{LR\&ALVM}~\cite{liu2020cross}; and \textbf{EN}~\cite{li2020mtop}.
The second group is the data transfer methods, including \textbf{EN+Trans.}~\cite{li2020mtop} and \textbf{EN+Trans.+mask}~\cite{li2020mtop}.

\subsection{Results}

\begin{table*}[h!]
\setlength{\abovecaptionskip}{3pt} 
\setlength{\belowcaptionskip}{-5pt}
\begin{center}
\small
\scalebox{0.9}{
 \begin{tabular}{l c c c c c c}
 \toprule
   & es & fr & de & hi & th & Average \\
 \midrule
   Classifier based selection 
   & 71.45 & 69.21 & 71.68 & 66.37 & 60.43 & 67.83 \\
   LM based selection
   & 70.41 & 68.78 & 70.95 & 65.94 & 58.44 & 66.90 \\
   \midrule
Classifier based selection  w/ relabeling
   & 72.01 & 70.06 & 73.29 & 69.01 & 60.25 & 68.92 \\
   LM based selection w/ relabeling
   & 71.38  & 69.28	 & 72.95 &	67.73 &	61.19 &	68.51 \\
 \midrule
  EN + Trans. + Gen. + Denoise
  & 71.58 & 71.53 & 74.05 & 69.66 & 62.31 & \textbf{69.83} \\
 \bottomrule
\end{tabular}
}
\end{center}
\caption{Comparison with other denoising methods on the MTOP dataset.}
\label{tab:other denoising}
\end{table*}

\begin{table*}[ht]
    \centering
    \small
    \scalebox{0.9}{
    \begin{tabular}{l c c c c c l}
     \toprule
      & es & fr & de & hi & th & Average \\
     \midrule
      EN + Trans.
      & 70.25 & 67.27 & 70.39 & 64.22 & 55.80 & 65.59 \\
      EN + Gen.
      & 66.71 & 65.99 & 70.84 & 63.55 & 45.32 & 62.48 \\
      EN + Trans. + Gen.
      & 69.21 & 66.93 & 71.46 & 65.48 & 56.89 & 65.99 \\
     \midrule
      EN + Trans. + Gen. + Denoise
      & 71.58 & 71.53 & 74.05 & 69.66 & 62.31 & \textbf{69.83} \\
      \quad w/o Gen.
      & 71.08 & 70.28 & 72.67 & 70.05 & 62.71 & 69.36$~\downarrow$ \\
     \quad w/o instance relabeling
     & 71.98 & 70.15 & 72.78 & 69.37 & 60.87 & 69.03$~\downarrow$  \\
     \quad w/o co-training
     & 71.71 &	69.56 &	73.15 &	68.94 &	 61.41 & 68.95$~\downarrow$\\
     \quad w/o instance re-weighting
     & 72.18 &	70.84 &	73.99 &	68.55 &	62.82 &	69.68$~\downarrow$  \\
     \bottomrule
    \end{tabular}
    }
    \caption{Ablation study on the MTOP dataset.}
    \label{tab:ablation study}
\end{table*}

\begin{figure}
\setlength{\abovecaptionskip}{3pt} 
\setlength{\belowcaptionskip}{-5pt}
    \centering
    \includegraphics[width=1.0\linewidth]{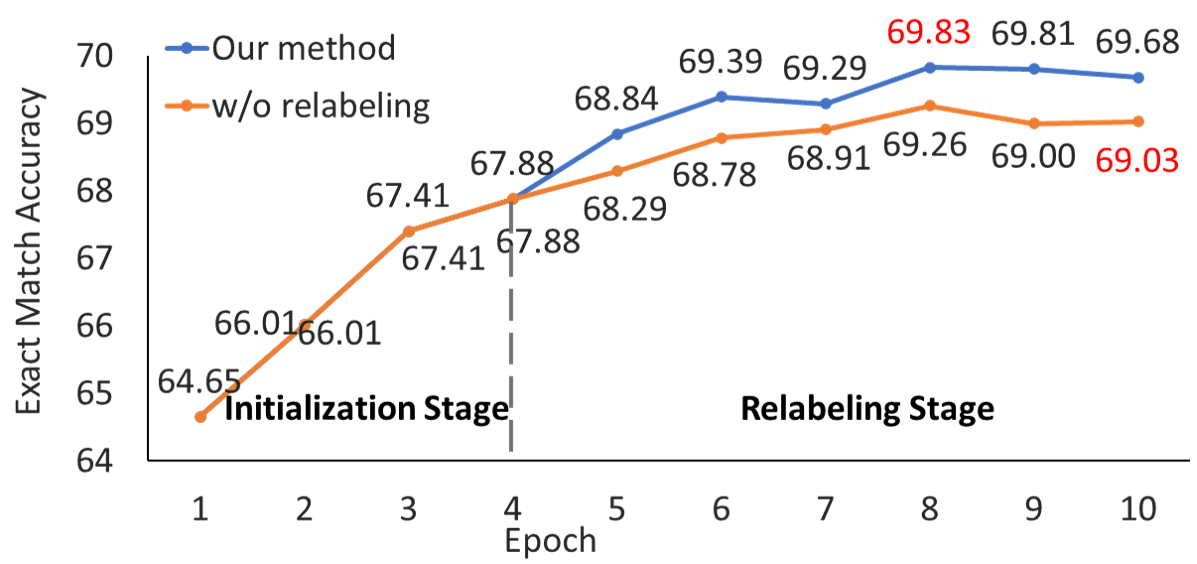}
    \caption{Metric trend per epoch on MTOP w/ or w/o instance relabeling strategy.}
    \label{fig:relabeling}
\end{figure}

Table~\ref{tab:main_table} reports the results of our approach
and the SOTA baselines. As the translator used in~\citet{li2020mtop} is not publicly available, we use Google translator instead, which leads to some results on some languages slightly different from reported by~\citet{li2020mtop}.

Our method outperforms the SOTA baselines and achieves new SOTA performance. Our method improves the Exact Match Accuracy on MTOP from 65.59 to 69.83, the F1 score on SNIPS from 81.44 to 84.49, and the accuracy on SNIPS from 98.64 to 98.71. These results clearly demonstrate the effectiveness of our proposed method. 

One interesting finding is that the performances on Spanish and French become slightly worse after adding the translated data. It is because the noise introduced by the machine translation and alignment processes may hurt the performance. Our method introduces the denoising training approach, which is able to handle the noise of synthesized data. 

\subsection{Comparison with Other Denoising Methods}
To verify the effectiveness of our approach, we conduct experiments with two kinds of denoising approaches used by the previous works. First, we consider the \textbf{classifier based selection} approach. Following \citet{anaby2020not}, we train an extra classifier using the corpus in English and the translated corpus, and filter out noisy data according to the probability scores predicted by the classifier. Second, we consider the \textbf{LM based selection} approach.  Following \citet{shakeri2020multilingual}, we use the language model score as the indicator to select high-quality data. 

For fair comparisons, for each noise filtering baseline, we also fine-tune two pre-trained XLM-R large models using different random seeds and take the ensembled model as the final model. Additionally, to remove the effect of instance relabeling method, we also apply it to baselines similar to our approach. Table~\ref{tab:other denoising} shows the comparison results on the MTOP dataset. Our approach outperforms those two methods (w/ relabeling) by 0.91 and 1.32 percentage points, respectively. This suggests that the gain of our method is not from the simple ensemble of two models. Instead, our method could indeed effectively remove the noise of synthesized data, outperforming previous noise filtering methods.   


\begin{table*}[t!]
\begin{center}
\resizebox{\textwidth}{!}{%
 \begin{tabular}{l c c c c c}
 \toprule
 Type & Source & Lang & Utterance & Original Labels & Modified Labels\\
 \midrule
 Intent & Gen. & fr & je suis tout disponible & get\_available & set\_available \\
 
 Intent & Gen. & de & verschieben sie die erinnerung in den alarm & update\_reminder & delete\_reminder \\
 
 Slot & Trans. & es & ¿cómo es el clima en \textcolor{red}{colorado} en esta época del año? & None & location \\
 
 Boundary & Trans. & es & resumen de noticias \textcolor{red}{[}de \textcolor{blue}{tayer}\textcolor{red}{]}& \textcolor{blue}{data\_time} & \textcolor{red}{data\_time} \\
  Slot\&Boundary & Gen. & de & lesen sie mir \textcolor{red}{[}die \textcolor{blue}{heutigen}\textcolor{red}{]} schlagzeilen & \textcolor{blue}{news\_source} & \textcolor{red}{data\_time} \\
 \bottomrule
\end{tabular}}
\end{center}
\caption{Cases of relabeling results on the MTOP dataset.}
\label{tab:case-study}
\end{table*}

\begin{table}[htbp]
\setlength{\abovecaptionskip}{3pt} \setlength{\belowcaptionskip}{-5pt}
\begin{center}
\resizebox{0.5\textwidth}{!}{
\small
 \begin{tabular}{l c c c}
 \toprule
 & One model & Two models  & Three models \\
 \midrule
 EN+Trans.+Gen.+Denoise & 67.57 & 69.83 & \textbf{70.04}\\
 \bottomrule
\end{tabular}
}
\end{center}
\caption{Performance of approaches with different number of models on the MTOP dataset.}
\label{tab:n-models}
\end{table}

\subsection{Ablation Study}

To validate the contribution of each component in our method, we conduct the ablation study on the MTOP dataset. We consider several ablation options. (1) \textbf{w/o generation} removes the generated training data. (2) \textbf{w/o instance relabeling} keeps the intent and slot labels of data unchanged throughout the training process.
(3) \textbf{w/o co-training} trains models using all training data without filtering.
(4) \textbf{w/o instance re-weighting} skips the instance re-weighting strategy. 

As shown in Table~\ref{tab:ablation study}, compared with the performance of approach using  translated and source language data, the performance of approach using generated data and source language data is reduced by 3.11 percentage points without denoising strategies. This is due to the much noise introduced by the generation process, which hurts the performance. When combining with our denoising approach, the approach with generated training corpus is superior to the approach without generated corpus by 0.47 percentage points. We consider that multiple augmented data sets increase data diversity and lead to better supervision signals.
Table~\ref{tab:ablation study} also shows that removing any of the other components generally leads to clear performance drop. It confirms that all of the proposed techniques contribute to the cross-lingual setting. 

\label{sec:analysis}

\subsection{Effect of Instance Relabeling} 

To better understand the effect of the instance relabeling strategy, in Figure~\ref{fig:relabeling}, we record the Exact Match Accuracy of our method with or without the relabeling strategy on the MTOP test set after each training epoch. The performance of our method with relabeling strategy keeps improving and is consistently better than the baseline during the relabeling stage. It demonstrates that the relabeling method indeed corrects many label errors in the noisy training data and the corrected labels contribute to the performance improvement.

\subsection{Effect of Number of Models} We explore the effect of the number $K$ of models (XLM-R). Specifically, we conduct experiments using one or three models. In the setting of one model, that is $K=1$, we only train one network with all training corpora $\mathcal{D}=\{\mathcal{D}_{src},\mathcal{D}_{trans},\mathcal{D}_{gen}\}$, and adopt the instance relabeling and instance filtering strategies. In the setting of $K=3$, three models are trained using $\{\mathcal{D}_{src},\mathcal{D}_{trans}\}$, $\{\mathcal{D}_{src},\mathcal{D}_{gen}\}$ and $\{\mathcal{D}_{src},\mathcal{D}_{trans},\mathcal{D}_{gen}\}$, respectively. 

The results shown in Table~\ref{tab:n-models} indicate that our method can effectively extend beyond two models. When the number of networks increases, the performance improves. The intuition is that more models can produce more reliable predictions, and thus can lead to better instance relabeling as well as instance filtering.

\begin{table}[t]
\setlength{\abovecaptionskip}{3pt} \setlength{\belowcaptionskip}{-5pt}
\begin{center}
\resizebox{0.5\textwidth}{!}{
\small
 \begin{tabular}{l c c c c c c}
 \toprule
 & 0\% & 10\% & 20\% & 30\% & 40\%\\
 \midrule
 EN+Trans.+Denoise & 68.32 & 68.69 & \textbf{69.36} & 69.07 & 68.75\\
 EN+Trans.+Gen.+Denoise & 68.95 & 68.77 & 69.12 & \textbf{69.83} & 69.07\\
 \bottomrule
\end{tabular}
}
\end{center}
\caption{Performance with different filtering rates and training corpus.}
\label{tab:noise-filtering}
\end{table}

\subsection{Effect of Filtering Rate in Co-Training}

To study the impact of the co-training strategy, we conduct experiments with different filtering rates on the MTOP dataset. Table~\ref{tab:noise-filtering} shows the results with regard to different filtering rates and different training corpora. For both approaches using or not using the generated corpus, as the filtering rate increases, the performance improves as well. This demonstrates that the filtering strategy can indeed filter out noisy instances effectively. However, further increasing the filtering rate degrades the performance. It is mainly because of the excessive drop of useful information contained in the training data. Another finding is that the best filtering rate for training corpus $\{\mathcal{D}_{src},\mathcal{D}_{trans},\mathcal{D}_{gen}\}$ is larger than that for $\{\mathcal{D}_{src},\mathcal{D}_{trans}\}$. The explanation may be that the generated corpus $\mathcal{D}_{gen}$ has more diverse data than the translated corpus $\mathcal{D}_{trans}$, but may also contain more noise.

\section{Case Study}
We conduct case analysis of the instance relabeling results on the MTOP dataset to examine the capability of our approach. We statistically analyze the differences between the original labels and the modified labels after relabeling stage.

We find that our instance relabeling method effectively corrects wrong labels of the synthesized data, including intent label and slot label as shown in Table~\ref{tab:case-study}. Specifically, there are four types of label modifications: 1) \emph{Intent Change}: the intent label of an utterance is modified; 2) \emph{Slot Change}: the slot type of a text span is modified; 3) \emph{Boundary Change}: the BIO boundaries of a slot are modified; and 4) \emph{Slot and Boundary Change}: both the slot type and the BIO boundaries are modified. For the MTOP dataset, intent labels of 4.99\% of the translated and generated data are modified and the slot labels of 33.10\% of those data sets are modified. 

From the case study, we can see that the synthesized data indeed contains much noise and our relabeling strategy is able to greatly reduce the negative impact of the noise by correcting different types of label errors. 


\section{Conclusions}
In this paper, we propose a denoising training approach where multiple models trained from various augmented methods provide supervision signals to each other. Extensive experimental results show that the proposed method outperforms the previous approaches, and can certainly alleviate the noisy label problem. Our proposed method is independent of the backbone network (e.g., XLM-R model) and the task. As future work, we plan to investigate the performance of our method on different cross-lingual tasks.

\section*{Acknowledgments}

Jian Pei's research is supported in part by the NSERC Discovery Grant program. All opinions, findings, conclusions and recommendations in this paper are those of the authors and do not necessarily reflect the views of the funding agencies.

\bibliography{anthology,custom}
\bibliographystyle{acl_natbib}
\clearpage
\appendix
\section{Experimental Setting}
\subsection{Datasets}
We evaluate the effectiveness of our proposed approach over five target languages on two benchmark datasets:
\textbf{SNIPS}~\citep{schuster2018cross} and  \textbf{MTOP}~\citep{li2020mtop}. Statistics of used data are detailed in Table~\ref{tab:data}.
 \begin{table}[h]
\centering
\scalebox{0.8}{
\begin{tabular}{lccc}
\hline
 & {\bf SNIPS} & {\bf MTOP}\\ \hline
 Language & en,es,th & en,es,fr,de,hi,th\\
 Intent types & 12 & 117 \\
 Slot types & 11 & 78 \\
 Train & 30,521 & 13,151 \\
 Dev & 4,181 & 1,878 \\
Test & 4,735 & 16,442 \\
 \hline
\end{tabular}}
\caption{Statistics of datasets.}
\label{tab:data}
\end{table}
\vspace{-20pt}
 
 \subsection{Implementation Details}
For generation, we fine-tune mBART pre-trained on 25 languages with 0.3 dropout, 0.2 label smoothing, 2500 warm-up steps, $3\times10^{-5}$ maximum learning rate, and 1024 tokens in each batch. For text filling, we mask $35\%$ of the words in each instance by randomly sampling a span length according to a Poisson distribution ($\lambda\!=\!3.5$). Then we append to each instance an end-of-sentence token ($<\!/S\!>$) and the corresponding language id symbol ($<\!LID\!>$). 
We don't search the best parameters for generation but use the default values in open-source code$\footnote{https://github.com/pytorch/fairseq/tree/master/examples\\/mbart}$. The final models are selected based on validation likelihood.

For SLU, we use XLM-R large model with about 550M parameters as the backbone network. In the fine-tuning process, we set the batch size as 128, fine-tuning epochs $E_{all}=10$, initialization epochs $E=4$ and 0.1 dropout for two benchmark datasets. The maximum filtering rates are $\delta=0.2$ and $\delta=0.3$ for the SNIPS and MTOP datasets, respectively. The learning rates are $2\times10^{-5}$ and $5\times10^{-5}$ for SNIPS and MTOP datasets, respectively. We select the best hyper-parameters by searching a combination of batch size, learning rate, the number of fine-tuning epochs, the number of initialization epochs and the filtering ratio with the following range: batch size $\{32,64,128\}$, learning rate $\{1,2,3,4,5\}\times10^{-5}$, fine-tuning epochs $\{5,10,15\}$, initialization epochs $\{2,3,4\}$, filtering ratio $\{10\%,20\%,30\%,40\%\}$. The models are saved by performance on the English development corpus and translated target language development corpus. The models are trained using mini-batch back-propagation, and the AdamW \citep{loshchilov2017decoupled} optimizer is used for optimization. We fine-tune the models on two V100-32GB GPUs which lasts about 4 hours.

\section{More Experimental Results and Discussions}

\subsection{Effect of Generated Data Size} To further analyze the effect of generated data, we randomly sample $a=\{0,1,2,3\}$ instances from the candidate set for each input to construct the generated corpus.
As shown in Table~\ref{tab:data-scale}, by increasing the size of the generated corpus, the performance improves. However, when the data size reaches a certain scale, e.g., $a=3$ in our experiments, the performance slightly regresses, but still outperforms the baseline without generated data. This suggests that the augmentation module indeed increases the diversity of training data and then improves the performance. Though increasing noise limits the growth of the improvement, our approach is robust enough to achieve comparable performance.

\begin{table}[t]
\setlength{\abovecaptionskip}{3pt} \setlength{\belowcaptionskip}{-5pt}
\begin{center}
\resizebox{0.5\textwidth}{!}{
\small
 \begin{tabular}{l c c c c}
 \toprule
 & 0 & 1 & 2 & 3\\
 \midrule
  EN+Trans.+Gen.+Denoise & 69.36 & 69.83 & \textbf{70.52} & 69.98 \\
 \bottomrule
\end{tabular}
}
\end{center}
\caption{Performance with different sizes of generated data on the MTOP dataset.}
\label{tab:data-scale}
\end{table}

\subsection{Variance Analysis}
We conduct 5 runs of training and calculate the mean and standard deviation (Stdev) values for our approach and the baseline on the MTOP dataset. The results are listed in Table~\ref{tab:t-test}. Besides, we also conduct a two-sided statistically significant t-test with the significance threshold 0.05 to compare the baseline with our method. The results show that the variance of our approach is similar to that of the baseline. Moreover, with p-value = $4.5\times10^{-8}$, our method outperforms the baseline with statistical significance.
\begin{table}[t]
\setlength{\abovecaptionskip}{3pt} \setlength{\belowcaptionskip}{-5pt}
\begin{center}
\resizebox{0.5\textwidth}{!}{
\small
 \begin{tabular}{l c c c c}
 \toprule
 & & Mean & Stdev \\
 \midrule
 Baseline & 65.59 66.02 65.59 65.46 65.12 & 65.56 & 0.29 \\
 Our method & 69.83 69.68 69.51 69.46 69.02 & 69.50 & 0.27 \\
 \bottomrule
\end{tabular}
}
\end{center} 
\caption{Mean and Standard deviation (Stdev) values across five different runs for our approach and the baseline on the MTOP dataset.}
\label{tab:t-test}
\end{table}

\end{document}